# Neuromorphic IoT Architecture for Efficient Water Management: A Smart Village Case Study

**Authors:**

Mugdim Bublin[1], Heimo Hirner[1], Antoine-Martin Lanners[1], Radu Grosu[2]

[1] University of Applied Science FH Campus Wien, Austria
[2] CPS, Technische Universität Wien (TU Wien), Austria

**Conference Topic:** Neuromorphic HW Architectures (digital, mixed signal, analog, spiking)

**Abstract**

The exponential growth of IoT networks necessitates a paradigm shift towards architectures that offer high flexibility and learning capabilities while maintaining low energy consumption, minimal communication overhead, and low latency. Traditional IoT systems, particularly when integrated with machine learning approaches, often suffer from high communication overhead and significant energy consumption.

This work addresses these challenges by proposing a neuromorphic architecture inspired by biological systems. To illustrate the practical application of our proposed architecture, we present a case study focusing on water management in the Carinthian community of Neuhaus. Preliminary results regarding water consumption prediction and anomaly detection in this community are presented. We also introduce a novel neuromorphic IoT architecture that integrates biological principles into the design of IoT systems. This architecture is specifically tailored for edge computing scenarios, where low power and high efficiency are crucial. Our approach leverages the inherent advantages of neuromorphic computing, such as asynchronous processing and event-driven communication, to create an IoT framework that is both energy-efficient and responsive. This case study demonstrates how the neuromorphic IoT architecture can be deployed in a real-world scenario, highlighting its benefits in terms of energy savings, reduced communication overhead, and improved system responsiveness.

# Introduction

The exponential growth of the Internet of Things (IoT) networks necessitates a paradigm shift in how these systems are designed and deployed. Traditional IoT systems, especially those integrated with machine learning techniques, often face significant challenges, including high communication overhead, increased energy consumption, and latency issues [1]. As IoT devices become more prevalent, particularly in edge computing scenarios, the need for architectures that balance flexibility, learning capability, and energy efficiency becomes critical. These issues are exacerbated when machine learning models are employed, as they typically require substantial computational resources. The challenge lies in creating an IoT architecture that can efficiently process data at the edge, with minimal energy consumption and latency, while still providing the flexibility and learning capabilities necessary for complex tasks such as anomaly detection and prediction.

This paper addresses these challenges by proposing a neuromorphic IoT architecture inspired by biological systems, specifically tailored for scenarios where low power consumption and high efficiency are essential. Neuromorphic computing is an innovative approach in computer engineering that designs computational systems inspired by the architecture and functionality of the human brain and nervous system [2-4]. This brain-inspired computing method offers significant advantages, including [5, 6]:

- **Energy Efficiency**: Traditional deep learning models may require up to 20 MW of power, while the human brain operates on just about 20 W, highlighting the potential for substantial energy savings.
- **Latency**: Neuromorphic systems excel in parallel processing, allowing for faster computations and reduced latency.
- **Safety & Security**: These systems enhance reliability by using redundant and analog components, mimicking the robustness of biological neural networks.
- **Reduced Costs** and **Waste**: By leveraging materials that mimic biological processes, neuromorphic computing can lower production costs and minimize environmental impact.

These benefits point towards a new computing paradigm that could revolutionize how we approach computational tasks. The primary objective of this study is to propose a neuromorphic IoT architecture that integrates principles from biological systems into the design of IoT systems. This architecture is intended to offer significant improvements in energy efficiency, communication overhead, and system responsiveness. Additionally, we aim to validate the effectiveness of this architecture through a case study on water management in Neuhaus, demonstrating its practical application and potential benefits.

## Neuromorphic IoT Architecture

### Design Principles

The proposed neuromorphic IoT architecture is inspired by the efficiency of biological systems, particularly human nervous system, which have evolved to process information with minimal energy consumption and high responsiveness [6]. The key features of this architecture include: distributed control and learning, prediction instead of commands and Reservoir computing.

### Distributed Control and Learning

Control and learning in an IoT network are distributed across various nodes, allowing the system to leverage locally available information for decision-making where it is most effective. This architecture is tailored for edge computing environments, prioritizing low power consumption and high efficiency. By processing data at the edge, the system can make real-time decisions without the need for constant communication with centralized servers.

In practice, different machine learning models are deployed across various layers of the IoT architecture: the device, or egde layer, fog layer, and cloud layer. At the edge layer, rapid decisions are made, such as automatically shutting off a water pipe in case of damage. The fog layer allows for more sophisticated decisions by integrating data from nearby devices to improve local water management. At the cloud layer, data from IoT devices and the internet is aggregated to manage overall water resources and address disaster scenarios.

We propose that higher architectural levels use **predictions instead of commands**, similar to the human visual [7] and motor systems [8]. In Fog and Cloud, models of water predictions on regional and global scales are built. The model predictions are sent to lower levels, while error corrections are sent from lower to upper layers. Unlike Federated Learning [9], which relies on a single global model for all devices, this approach utilizes local machine learning models to make decisions based on locally available information. This strategy reduces communication overhead, latency, and energy consumption.

**Neuromorphic Reservoir Computing** is particularly well-suited for edge implementation due to its low computational overhead, which results in reduced energy consumption, and its ability to harness underlying physical processes for computation. Neuromorphic Reservoir Computing is a computational framework inspired by the brain's neural networks, particularly suited for processing time-series data and performing complex pattern recognition tasks [10-12]. This approach is based on the concept of a "reservoir," which is a recurrent neural network with fixed, randomly initialized connections. The key idea is that the reservoir can transform temporal input data into a higher-dimensional space, where the information is easier to analyze and predict. The advantage of this method lies in its simplicity: only the output layer is trained, while the reservoir itself remains unchanged. This reduces the computational burden associated with training deep neural networks.

In Figure 1 the analogy between human nervous system and the proposed IoT architecture is depicted.

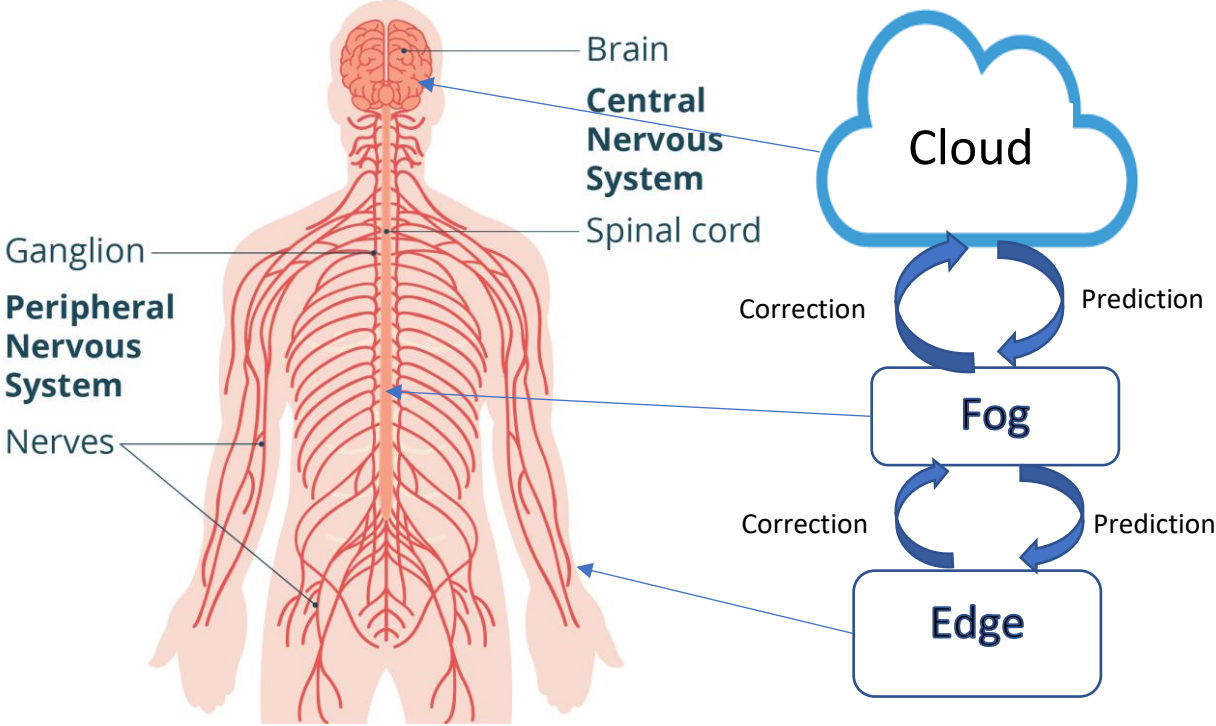

*Figure 1: Analogies between human nervous system and the proposed IoT architecture*

**Free Energy Principle**

Friston's Free Energy Principle [13], [14] is a theoretical framework from neuroscience, which posits that the brain minimizes a quantity called free energy to maintain a stable internal state and make sense of the world. This principle is derived from thermodynamics and statistical mechanics, and it explains how biological systems (like the brain) resist disorder (entropy) by maintaining an internal model of the environment. In our case, each level of the IoT hierarchy has its own world model that is updated based on inputs from the next lower level.

The free energy principle can also be broken down into terms involving surprise and approximation errors.

$$F = DKL(q(s) \parallel p(s \mid o)) - \log p(o)$$

Where:

$DKL(q(s) \parallel p(s|o))$ is the Kullback-Leibler (KL) divergence between the recognition density $q(s)$ and the posterior distribution $p(s \mid o)$, which measures how different the brain's internal model is from the real posterior probability of hidden states given observations.

$\log p(o)$ is the log-evidence or surprise associated with the sensory input $o$. The brain aims to minimize surprise (unexpected sensory states).

To reduce free energy, the brain does two things:

**Perception**: Adjusting its internal model (recognition density $q(s)$ to match sensory input better. This helps reduce the Kullback-Leibler divergence (KLD), which improves the accuracy of its beliefs about hidden states.

**Action**: Taking actions to influence the environment in a way that makes sensory input more predictable, thus reducing surprise $\log p(o)$.

In our setup, each hierarchy level of an IoT network creates its own internal model of the environment, updates the model according to sensor inputs from lower layers, and performs actions on the lower layers to reduce discrepancies between the model and the environment.

The free energy principle enables **more autonomy** for each layer, as it places fewer constraints on the internal model of each layer compared to, for example, reinforcement learning (RL), which assumes that agents maximize expected rewards. Furthermore, it also requires **less communication overhead**, since information is signaled to other layers only when the differences between predictions and actual measurements exceed a certain threshold.

**Asynchronous Processing and Event-driven Communication**

Unlike traditional computing systems, which rely on a global clock, the proposed architecture processes information asynchronously. This allows for more efficient use of resources, as computation only occurs when necessary.

In biological systems, communication between neurons occurs only when a certain threshold is reached, triggering a spike of activity. This principle is applied in the proposed architecture to reduce unnecessary communication, thereby lowering energy consumption and latency.

## The Role of Thresholds in Hierarchical IoT Model

In hierarchical models, each layer generates predictions based on the lower layer's output, with higher layers handling more abstract representations. Thresholds, in this context, can refer to parameters that decide:

- **Uncertainty bounds**: Determining when prediction errors (discrepancies between predicted and observed states) should trigger adjustments in the model or action.
- **Signal passing thresholds**: Deciding when sufficient confidence is achieved in a layer to propagate the information upward or downward.

In active inference framework agents (IoT layers) continuously minimize their **variational free energy** (a measure of surprise or uncertainty about sensory inputs) by updating beliefs or performing actions. Key to this process is minimizing **prediction errors** through updating the generative model (beliefs about the world) or by acting to bring sensory inputs in line with predictions.

- **Perception**: Update internal states to minimize prediction error.
- **Action**: Perform actions to reduce the discrepancy between expected and actual sensory inputs.

**Setting Adaptive Thresholds Using Prediction Errors**

In active inference, **prediction error** (the difference between the predicted input and actual sensory input) drives updates to beliefs or actions. To adaptively set thresholds at different hierarchical layers, you can allow thresholds to be modulated by the **magnitude of prediction error** and the **precision (inverse uncertainty)** associated with each layer's predictions.

1. **Prediction Error Calculation**: At each layer $L_i$, prediction error is computed as:

$$\epsilon_i = input_i - prediction_i$$

where $\epsilon_i$ is the prediction error at layer $i$, and $input_i$ is the input to the layer (could be sensory data for the bottom layer or output from the previous layer).

2. **Precision-Weighted Prediction Error**: To adaptively set thresholds, weight the prediction error based on the layer's **precision** $\Pi_i$ which represents the inverse of uncertainty:

$$\tilde{\epsilon}_i = \Pi_i \cdot \epsilon_i$$

Here, $\tilde{\epsilon}_i$ is the precision-weighted prediction error.

3. **Adaptive Threshold Adjustment**: Let the threshold at layer $L_i$ denoted $\tau_i$ be adjusted based on the running average or variance of prediction errors at that layer. For example, using an exponentially weighted moving average (EWMA):

$$\tau_i(t+1) = \alpha \cdot \tau_i(t) + (1-\alpha) \cdot |\tilde{\epsilon}_i|$$

where $\alpha$ is a smoothing factor, controlling how quickly the threshold adapts to changes in prediction error magnitude.

4. **Hierarchical Precision Tuning**: Active inference frameworks often adjust precision at each layer based on uncertainty in the environment. If a lower-level layer exhibits high variability (uncertainty), precision at higher levels may be reduced (lower confidence in predictions), increasing tolerance for prediction errors.

   We can compute adaptive precision $\Pi_i$ at each layer based on past errors:

   $$\Pi_i = \frac{1}{\sigma_i^2 + \beta}$$

   where $\sigma_i^2$ is the variance of prediction errors at layer $L_i$, and $\beta$ is a small constant to avoid division by zero.

5. **Threshold Propagation Through Layers**: In a hierarchical model, the adaptive thresholds $\tau_i$ can also be influenced by errors at neighboring layers. For instance:

   $$\tau_i = f(\epsilon_{i-1}, \epsilon_{i+1})$$

   where $\tau_i$ is set adaptively based on errors in both the layer above and the layer below, helping each layer balance local and global model adjustments.

**Incorporating Actions into Threshold Setting**

Active inference involves both **belief updates** and **action selection**. Thresholds can be adapted based on both the **perceptual** prediction errors (used to update internal beliefs) and the **action** outcomes (used to reduce discrepancy between predicted and actual sensory inputs).

For example:

- If actions reduce prediction error, thresholds may decrease, indicating higher precision in predictions.
- If actions increase prediction error, thresholds may increase to allow more flexibility in updating the generative model.

**Optimizing Thresholds Using Free Energy Minimization**

In active inference, the agent seeks to minimize **free energy**, which combines **prediction error** and **model uncertainty**. Thus, thresholds $\tau_i$ can be adaptively set by minimizing the total free energy at each hierarchical level.

The free energy at each layer i can be expressed as:

$$F_i = \frac{1}{2}\tilde{\epsilon}_i x^2 + H(\Pi_i)$$

where $H(\Pi_i)$ represents the entropy or uncertainty in the precision at that layer. Minimizing $F_i$ can help set optimal thresholds at each layer by balancing prediction accuracy and uncertainty.

**Threshold Setting Summary**

To adaptively set thresholds in hierarchical layers using active inference:

1. **Monitor prediction errors** at each layer.
2. **Adjust thresholds** based on the precision-weighted errors.
3. **Tune precision** and thresholds across layers by minimizing free energy and propagating uncertainty estimates.
4. **Use feedback from action** to refine thresholds.

This approach aligns with the active inference principle of reducing prediction errors while accounting for uncertainty in a dynamic environment.

# Implementation

To implement the proposed architecture, we developed a neuromorphic IoT framework that integrates neuromorphic devices for data processing. These devices are deployed on edge devices, which are responsible for collecting data, processing it locally, and making decisions in real-time. The framework is designed to be modular, allowing for easy integration with existing IoT systems and flexibility in terms of the types of sensors and devices used.

# Case Study: Smart Village Water Management

## Context and Objectives

The municipality of Neuhaus (Suha) in south-eastern Carinthia on the Slovenian border with 1015 inhabitants (as of 1 January 2024, https://www.statistik.at) and an area of 36.34 km² offers a unique opportunity to apply the proposed neuromorphic IoT architecture in a real-life scenario. Like other similarly structured micro-communities, Neuhaus is facing the typical problems of rural areas in Central Europe: declining population figures combined with shrinking financial and human resources and new challenges due to climate change. Water management is of crucial importance in this region, as efficient utilisation of water resources is essential for both environmental sustainability and economic viability. In the last two years, Neuhaus was confronted with a period of drought in the summer of 2023 and enormous amounts of rain, flooding, and landslides in August 2024. Both have a negative impact on financial resources and the quality of life in the region.

In order to overcome these challenges and reduce the administrative expenses, the municipality decided back in 2020 to replace all 377 water meters in the municipality with smart meters. The water meters have to be replaced every 4 years anyway for calibration reasons, so the one-off additional costs were limited. A municipal LoRaWAN radio network was set up for data transmission. LoRaWAN fulfils the required criteria for long range in rural areas, low energy consumption of battery-operated smart meters, and low installation and operating costs. The first automatic meter reading took place in September 2021. In addition to the water meters, the reservoirs of the three separate water supply systems, in which the spring water is collected, were equipped with solar-powered level and flow meters. A total of around 560 LoRaWAN sensors are currently installed in the municipal area. In addition to the water meters, these include weather stations, road temperature sensors, snow height gauges, and indoor $CO_2$ sensors. In July 2022, the research cooperation between the municipality of Neuhaus and our University of Applied Sciences FH Campus Wien was launched. The aim of this cooperation is to do research on IoT solutions for small municipalities. This gives us access to all measurement data of the LoRaWAN sensor network. Instead of simulated laboratory data, we can work with real data, coming from a lossy IoT network. The first project realised as part of this collaboration was a water management system for real-time water balances tailored to small communities with their limited resources. It has been used by the community since July 2023.

In the present case study, we extend this system to include predictions and alerts. The problems we are facing are typical for IoT sensor networks. We have limited edge devices (our battery-powered water meters) and an unreliable network with very limited data rates. Typically, the transmission of water meter data occurs once daily. Only in the event of a pipe break after the meter should an immediate alarm message be sent. So, the decision must be made locally at the edge device. And, since these devices are battery-powered, they are also limited in their processing power and memory. The objectives of this case study are to:

1. Predict water consumption patterns in the community.
2. Detect anomalies in water usage that may indicate leaks or other issues.
3. Demonstrate the effectiveness of the proposed architecture in reducing energy consumption, communication overhead, and latency.

## Data Collection and Preprocessing

Data on water consumption was collected from various sensors deployed throughout the community. This data included hourly and daily water usage statistics, from the smart meters, as well as information on environmental factors such as temperature and humidity, which may influence water consumption patterns. The data was preprocessed to remove noise and fill in missing values due to losses in the LoRaWAN network before being fed into the neuromorphic IoT framework.

The following table shows the input and output information available at each IoT layer.

*Table 1: Input and output information available at each IoT layer*

| IoT Layer | Input from layer above | Other Inputs | Output to above layer | Time and space scales |
|---|---|---|---|---|
| Edge | Local water consumption threshold | Actual water consumption of each household | Difference between expected average local water consumption and the threshold | Seconds/Local (household) |
| Fog | Regional water consumption threshold | Regional water supply, Regional water | Difference between expected average regional | Days/Regional |

|   |   | pipe connection map | water consumption and the threshold |   |
|---|---|---|---|---|
| Cloud | - | Time, date and season<br><br>Weather information from internet<br><br>Water supply | Output to human: statistics, predictions and alarms over global water consumption | Months/Global |

## Prediction Models and Performance

To evaluate the effectiveness of the proposed architecture, we compared the performance of several machine learning models in predicting water consumption. The models used included a Multi-Layer Perceptron (MLP), Long Short-Term Memory (LSTM) network, AutoRegressive Integrated Moving Average (ARIMA) model, and Random Forest. The mean absolute percentage error (MAPE) was calculated for each model to assess their predictive accuracy (see Table 2).

*Table 2: Hourly and daily prediction errors of different algorithms*

| Algorithm | Hourly Prediction MAPE [%] | Daily Prediction MAPE [%] |
|---|---|---|
| MLP | 41.10 | 5.05 |
| LSTM | 33.51 | 4.87 |
| ARIMA | 30.06 | 5.18 |
| Random Forest | 26.05 | 5.40 |

The results indicate that the Random Forest model performed the best in terms of hourly prediction accuracy, with a MAPE of 26.05%. For daily predictions, the LSTM model achieved the lowest MAPE of 4.87%.

Overall, the ARIMA model achieves quite good performance in both hourly and daily water consumption prediction.

ARIMA (AutoRegressive Integrated Moving Average) is a classical model used for time-series forecasting, which combines autoregression, differencing, and moving average components.

Reservoir computing can approximate ARIMA by leveraging its ability to model nonlinear relationships and memory of past inputs, which are essential components of ARIMA models. In particular, the recurrent nature of the reservoir allows it to capture the autoregressive and moving average aspects of the time series, while the nonlinear transformation within the reservoir can approximate the differencing and other complex relationships present in ARIMA models. The **equivalence of reservoir computing to nonlinear vector autoregression (NVAR)** is rooted in the way both models handle temporal data [15], [16].

Nonlinear Vector Autoregression is a statistical model that predicts future values of a time series based on past values, accounting for possible nonlinear relationships between variables. Reservoir computing, by transforming input sequences into a dynamic state within the reservoir, effectively performs a similar operation to NVAR. It creates a nonlinear mapping of past inputs, which can then be used to predict future values. This equivalence is particularly useful because it shows that reservoir computing can be viewed as a form of nonlinear autoregression, with the reservoir acting as the nonlinear transformation that enables the prediction of future states based on past data. This understanding opens up the possibility of using reservoir computing as a powerful tool for time-series prediction, where traditional methods like NVAR are used.

In summary, by exploiting the equivalence between reservoir computing and nonlinear vector autoregression, reservoir computing can be effectively used to approximate the behavior of ARIMA models, offering a powerful alternative for time-series forecasting that is particularly well-suited to handling nonlinear and complex patterns in the data. Neuromorphic Reservoir Computing is particularly interesting for IoT networks due to its low computational overhead and wide range of possibilities for physical implementations.

### Anomaly Detection

Anomaly detection is a critical aspect of water management, as it allows for the identification of unusual patterns in water usage that may indicate leaks or other issues. In this study, we applied both *the mean + 3 sigma* rule and an LSTM-based anomaly detection method to the water consumption data. The results showed that the mean + 3 sigma rule detected more anomalies than the LSTM model, suggesting that simple statistical methods may be more effective for this type of application in certain contexts.

## Discussion

### Energy Efficiency and Communication Overhead

One of the primary advantages of the proposed neuromorphic IoT architecture is its energy efficiency. By processing data locally at the edge and using event-driven communication, the system significantly reduces the need for constant data transmission to centralized servers, thereby lowering energy consumption and communication overhead. This is particularly important in rural areas like Neuhaus, where energy resources may be limited.

### System Responsiveness and Latency

The asynchronous processing and event-driven communication of the proposed architecture also contribute to improved system responsiveness. In real-time applications such as water management, where timely detection of anomalies is crucial, the ability to process data and make decisions quickly can prevent significant water loss and reduce the environmental impact.

### Safety & Security

**Safety:** For analog systems it is possible to use continuity properties when pondering system behavior in different points of their state space. If a system exhibits intended behavior in a state A and in a related state B, it can be argued that it will show intended behavior also when

$C=\alpha A+(1-\alpha)B$, where $0<\alpha<1$. So if the system is tested in states A and B, then it can be assumed that it will not change too much in the intermediate states C in between.

**Security**: It is more difficult to access and modify analog hardware like memristors or reservoirs than to perform a security attack over the internet.

**Practical Implications**

The case study demonstrates that the proposed neuromorphic IoT architecture is not only theoretically sound but also practically viable. The deployment in Neuhaus and preliminary results show that the architecture can handle the complexities of a real-world environment while delivering tangible benefits in terms of energy savings, reduced communication overhead, and improved system responsiveness.

# Conclusion

The exponential growth of IoT networks demands a new approach to system architecture, one that balances flexibility, learning capability, and energy efficiency. This paper has proposed a neuromorphic IoT architecture inspired by biological systems, designed to meet these challenges in edge computing scenarios. Through a case study on water management in the Carinthian community of Neuhaus, we have demonstrated the practical application and benefits of this architecture. The results show that the proposed architecture can deliver significant improvements in energy efficiency, communication overhead, and system responsiveness, making it a promising solution for the future of IoT networks.

**Future Work**

While the proposed architecture has shown promising results, further research is needed to optimize the integration of neuromorphic computing with existing IoT frameworks and to explore its application in other domains. Additionally, the development of more sophisticated anomaly detection methods, potentially integrating neuromorphic principles, could further enhance the system's capabilities. As IoT networks continue to evolve, the principles outlined in this paper will be crucial in guiding the development of next-generation systems that are both efficient and effective